\documentclass[lettersize,journal]{IEEEtran}
\usepackage{amsmath,amsfonts}
\usepackage{algorithmic}
\usepackage{algorithm}
\usepackage{array}
\usepackage[caption=false,font=normalsize,labelfont=sf,textfont=sf]{subfig}
\usepackage{textcomp}
\usepackage{stfloats}
\usepackage{url}
\usepackage{verbatim}
\usepackage{graphicx}
\usepackage{cite}
\usepackage[colorlinks=false,linkcolor=black,urlcolor=black]{hyperref}
\usepackage{comment}
\usepackage{color}

\usepackage{xcolor, soul}

\begin{document}

\bstctlcite{IEEEexample:BSTcontrol}

\title{Integrating 3D Slicer with a Dynamic Simulator for Situational Aware Robotic Interventions}

\author{Manish Sahu$^{1}$, Hisashi Ishida$^{1\dagger}$, Laura Connolly$^{1, 3\dagger}$, Hongyi Fan$^{1\dagger}$, Anton Deguet$^1$, Peter Kazanzides$^1$, Francis X. Creighton$^{1, 2}$, and Russell H. Taylor$^{1, 2}$,~\IEEEmembership{Life Fellow,~IEEE}, Adnan Munawar$^{1}$ 
\thanks{$^1$Department of Computer Science, Johns Hopkins University, Baltimore, MD, USA. $^2$ Department of Otolaryngology-Head and Neck Surgery, Johns Hopkins University School of Medicine, Baltimore, MD, USA.
$^3$ School of Computing, Queen’s University, Kingston, ON, Canada }
\thanks{$\dagger$ These authors contributed equally. Email: {\tt manish.sahu@jhu.edu}}%
\thanks{*Presented as a podium presentation at the HSMR}
}

\markboth{Journal of \LaTeX\ Class Files,~Vol.~14, No.~8, August~2023}%
{Shell \MakeLowercase{\textit{et al.}}: A Sample Article Using IEEEtran.cls for IEEE Journals}


\maketitle
\begin{abstract}

Image-guided robotic interventions represent a transformative frontier in surgery, blending advanced imaging and robotics for improved precision and outcomes. This paper addresses the critical need for integrating open-source platforms to enhance situational awareness in image-guided robotic research. 
We present an open-source toolset that seamlessly combines a physics-based constraint formulation framework, AMBF, with a state-of-the-art imaging platform application, 3D Slicer.
Our toolset facilitates the creation of highly customizable interactive digital twins, that incorporates processing and visualization of medical imaging, robot kinematics, and scene dynamics for real-time robot control.
Through a feasibility study, we showcase real-time synchronization of a physical robotic interventional environment in both 3D Slicer and AMBF, highlighting low-latency updates and improved visualization.

\end{abstract}


\section{Introduction}

Image-guided robotic interventions represent an evolving frontier in surgical procedures, seamlessly blending robotic and imaging technologies to equip surgeons with enhanced information, refine precision, and improve outcomes \cite{fichtinger2022image}. Active imaging serves as a navigation tool for guiding surgical interventions~\cite{cleary2010image}, while robotic assistance enhances proficiency by refining tool tip precision and mitigating the impact of hand tremors~\cite{taylor1999steady}.

The development of intelligent image-guided robotic systems necessitates the integration of complementary situational awareness~\cite{simaan2015intelligent}, incorporating contextual models of patients and intra-operative devices. These models interpret surgical situations, delivering real-time feedback tailored to the ongoing procedure. Physics-based robotics simulations facilitate dynamic interaction with robotic models in a surgical virtual environment~\cite{choi2021use}. Presently, 3D Slicer~\cite{fedorov20123dSlicer}, an open-source medical imaging platform, stands as a widely adopted tool for research and prototyping in image guidance~\cite{ungi2016open}. Simultaneously, Robot Operating System (ROS)~\cite{quigley2009ros} is a widely popular open-source communication middleware for robotics research. However, 3D Slicer lacks a straightforward interface for surgical robotics applications~\cite{connolly2021open}. To fully leverage the potential of image-guided robotic surgery, there is a crucial need to integrate open-source navigation and robotics platforms.

Developing complementary situational awareness for image-guided robot-assisted research encounters a significant challenge: the absence of flexible open-source frameworks that empower researchers to swiftly develop and test systems tailored to specific surgical contexts. The simulation requirements for a situation-aware system are multi-faceted, encompassing the need for a real-time, physics-based dynamic virtual environment. This environment should seamlessly integrate various components, including surgical robots and patient anatomy, and extend its capabilities to provide real-time feedback aligned with safety constraints and the surgical context. Moreover, to ensure wide robotics compatibility, the platform must accommodate any robot supported in ROS.

This paper outlines the system development work focused on integrating 3D Slicer and a physics-based simulation environment, Asynchronous Multibody Framework (AMBF)~\cite{ambf}, using ROS as the underlying communication middleware. The objective is to empower researchers by merging the visualization and registration capabilities of 3D Slicer with the dynamic and constraint-based simulation capabilities of AMBF. Through this integration, researchers gain a comprehensive toolset to advance projects in medical robotics, contributing to the overall progression of the field.

To facilitate seamless integration, ROS modules were developed for both AMBF and 3D Slicer, functioning as ROS nodes. These modules act as a bridge, conveying contextual information between 3D Slicer and AMBF. This collaborative approach leverages the strengths of both tools, with 3D Slicer providing calibration, registration, and navigation visualization, and AMBF delivering real-time physics-based constraints. The incorporation of ROS not only facilitates communication but also opens the door to potentially integrate other open-source projects. To showcase the capabilities of our integrated system, we conducted a feasibility study, demonstrating low-latency, real-time visualization of robots within the 3D Slicer environment.

\begin{figure*}[htbp]
  \centering
  {\includegraphics[width=0.75\textwidth]{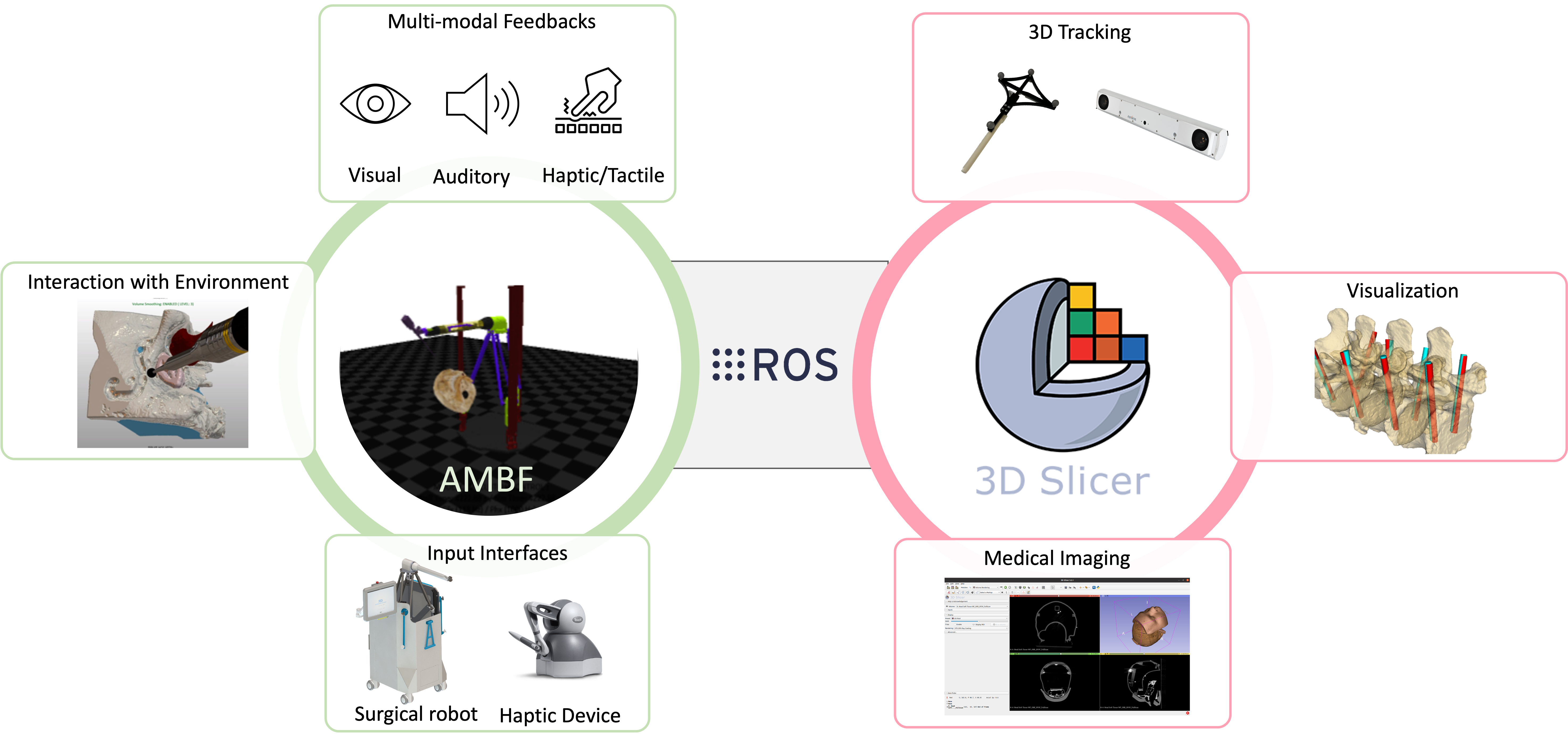}}
  \caption{Overview of integration of 3D Slicer and AMBF. Different features for each software tool are highlighted around their respective logos.}
\end{figure*}

\section{Related Work}

\textbf{Image-guidance in medical robotics}: 3D Slicer~\cite{fedorov20123dSlicer} is the most widely used medical imaging platform for research and prototyping, offering robust functionalities such as segmentation, registration, and three-dimensional visualization of medical image data.
Its dedicated image-guided therapy extension, SlicerIGT~\cite{slicerIGT}, further enriches its capabilities by providing image registration and interfacing with external devices to support image guidance.
An open-source communication protocol, OpenIGTLink~\cite{tokuda2009openigtlink}, further extends its utility for IGT research by providing a direct interface with commercial tracking systems\cite{lasso2014plus}.
To foster image-guided robotics research, there have been attempts to integrate 3D Slicer and ROS by using a ROS interface with OpenIGTLink~\cite{tauscher2015openigtlink} or a ROS-IGTL bridge~\cite{frank2017ros} implemented for the KUKA lightweight robot (LWR).
However, these interfaces merely facilitate low-level data exchange between the two environments~\cite{ros2_module_connolly}.
To enable more synchronous and seamless integration between the two platforms, Connolly et al.~\cite{ros2_module_connolly} developed a SlicerROS2 module to facilitate bi-directional data transfer between the Slicer Medical Reality Modelling Language (MRML~\cite{fedorov20123dSlicer}) scene to the ROS scene graph (Tf~\cite{foote2013tf}).

\textbf{Dynamic Simulation}: Within the ROS community, RViz is the default visualization tool, however, it does not simulate robot kinematics, dynamics or interaction. While several open-source simulators such as Gazebo \cite{koenig2004gazebo} and Virtual Robot Experimentation Platform (V-REP) \cite{rohmer2013vrep} support features for robot dynamics, these are not suitable for real-time surgical applications\cite{ambf}.
Asynchronous Multi-Body Framework (AMBF)~\cite{ambf}, which enables the simulation of complex surgical robots and environments, is modular, real-time, and asynchronous, offering direct access to software- (e.\,g., system libraries or drivers) and hardware components (input and haptic devices, optical tracking, video input, etc.).
Additionally, AMBF provides a hierarchical software plugin interface for the development of customized applications. Leveraging these design choices and capabilities, AMBF has been successfully expanded into an immersive simulation system for surgical purposes~\cite{munawar2023fully,phalen2023platform,shu2023twin,munawar2022open}.

In this work, we aim to integrate 3D Slicer and AMBF using Robot Operating System (ROS) to utilize the extensive capabilities of 3D Slicer for visualization, processing, and registration of medical imaging data, and the physics-based constraints of AMBF for simulating the interaction of a robot with the anatomy.

\section{Methods}


We provide the system overview and functionality of our toolset in this section.

\subsection{System Design}
The functional requirements for prototyping of image-guided robotic systems necessitate real-time visualization of the surgical scene in 3D Slicer, coupled with bidirectional data transfer between the simulated scene and the Slicer MRML scene.
The system design addresses these requirements by representing the scene in both AMBF and 3D Slicer. A crucial component of the design is providing communication between the MRML scene in 3D Slicer and the scene in AMBF represented by AMBF Description Format (ADF)~\cite{munawar2023fully}). To this end, we developed two modules - \emph{AMBF-ROS module} and \emph{Slicer-ROS module}. These modules contribute to the system by providing a high-level real-time description of the environment.

\begin{figure}[!htp]
  \includegraphics[width=0.47\textwidth]{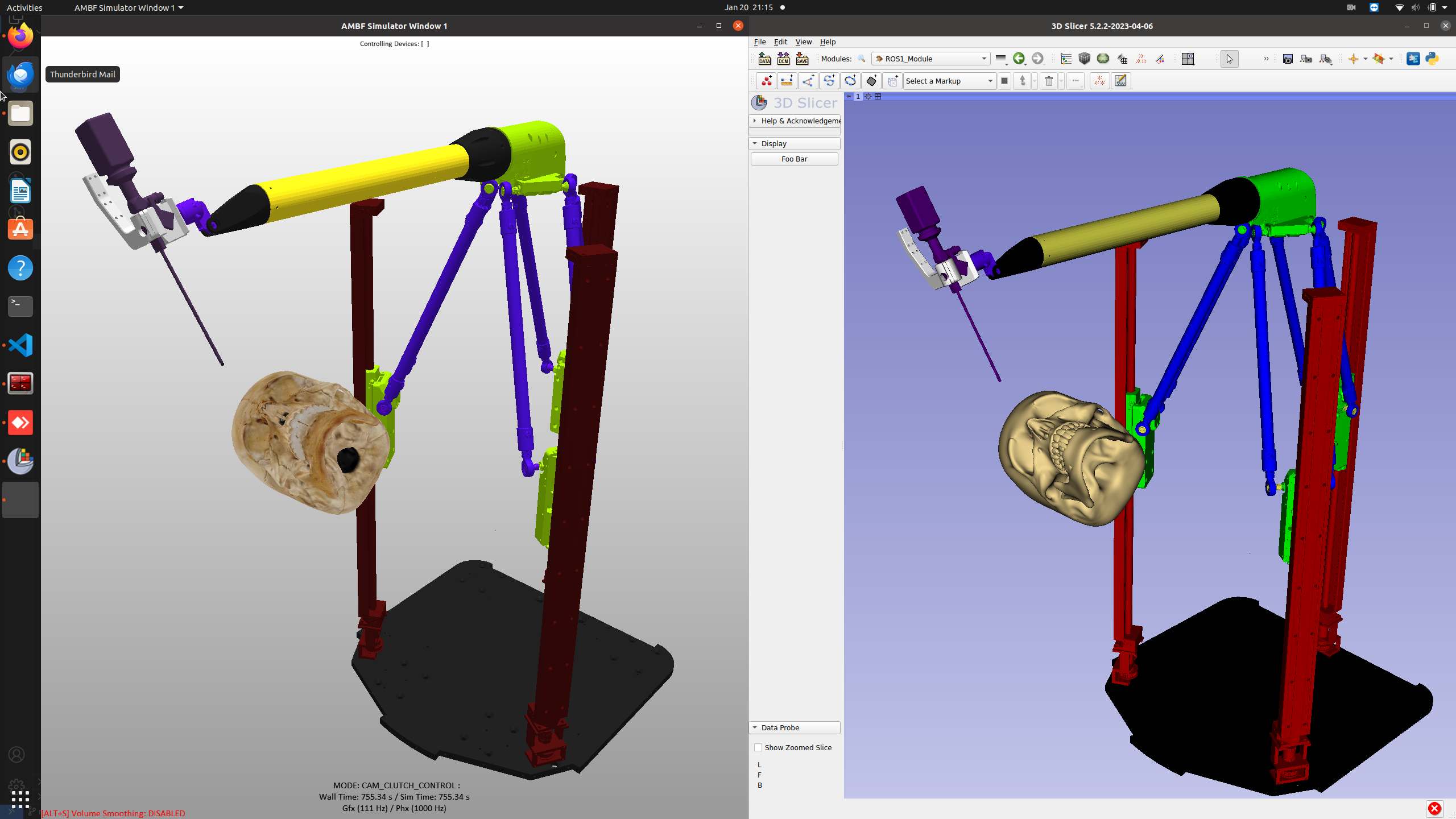}
  \centering
  \caption{Simulated robotic sinus endoscopy in AMBF (left), and real-time synchronized scene in 3D Slicer (right). The physics interaction between the robot and the anatomy in AMBF will be reflected in 3D Slicer.}
  \label{fig:model}
\end{figure}

\subsection{Scene Representation and Updates}

\subsubsection{AMBF}

In comparison to prior work~\cite{ros2_module_connolly}, our system provides support for both URDF and a customized message structure representing scene data, ADF, encompassing body names (rigid bodies) and system filepaths for meshes. This adaptation became essential for the AMBF simulator, which handles relative poses of rigid bodies at a lower level, allowing direct queries of rigid body poses through the ADF file.  During initialization, the simulator generates a custom message stored on the ROS parameter server. Subsequently, our 3D Slicer extension parses this message to seamlessly load the robot's rigid bodies.

\begin{figure}[!tp]
  \includegraphics[width=0.47\textwidth]{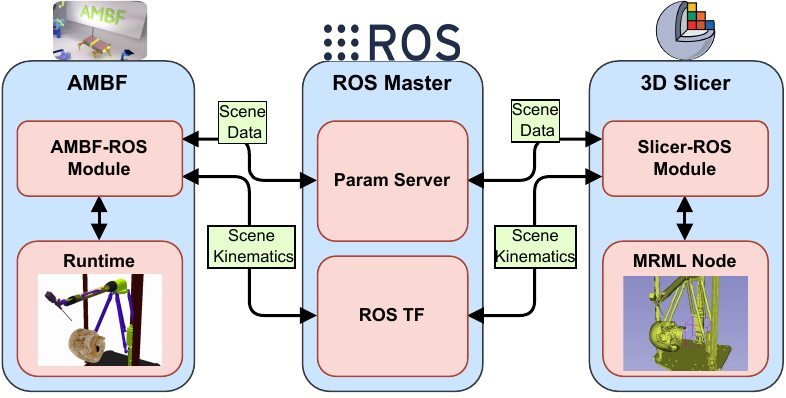}
  \centering
  \caption{Higher level structure of Slicer-ROS Module connected with AMBF Simulator. AMBF simulator and 3D slicer communicate Scene data and kinematics using ROS as an intermediary middleware.}
  \label{fig:model}
\end{figure}


\subsubsection{3D Slicer}
The 3D Slicer scene is organized according to MRML nodes that correspond to different data types. For this work, the links of the robot are represented as \emph{vtkMRMLModelNodes} and their respective positions are stored in \emph{vtkMRMLLinearTransformationNodes}. In addition to the robot, 3D Slicer natively offers several visualization tools for different medical imaging formats including ultrasound, computed tomography (CT) and magnetic resonance imaging (MRI). One of the main utilities of 3D Slicer is that this imaging information can be registered to other data in the scene using the SlicerIGT modules \cite{ungi2016open}. 


\noindent \textbf{Updates}: The 3D Slicer ROS module utilizes the ROS Transformation package (\textbf{tf}) to obtain the robot's state. The AMBF simulator plugin updates each rigid body's position through tf, allowing the 3D Slicer's ROS module to query each rigid body's transformation relative to the world frame and update the corresponding models within the MRML scene. To synchronize the 3D viewer in Slicer with the simulator, a custom timer (QTimer from the Qt library) was implemented in the 3D Slicer ROS module. This timer refreshes at $200 Hz$, triggers the ROS callbacks, and updates the MRML scene.


\subsection{System dependencies and compatibility}

To successfully build the system and replicate the results presented in this article, the following dependencies must be satisfied:

\begin{itemize}
    \item \textbf{3D Slicer}: It is crucial to have 3D Slicer built from source since the extension was developed within the 3D Slicer platform.
    \item \textbf{ROS Noetic}: The ROS1 communication infrastructure was utilized for data transmission between various components. Therefore, the installation of ROS Noetic, along with its standard packages, is necessary. 
    \item \textbf{AMBF and Plugin}: AMBF and the plugin designed for the task must be built from source. The AMBF plugin plays a key role in constructing the scene simulation and updating its state within the 3D Slicer environment.
\end{itemize}

Our system can be integrated with other ROS packages provided they use \textit{URDF} for robot description and \textit{tf} for scene updates.

\section{Experiments}


The primary focus of our work is to evaluate the real-time behavior of the system in a surgical robotics use case, specifically assessing communication performance for surgical scene updates.

\subsection{Surgical robotics use case}
To demonstrate the applicability of our system and assess its efficacy, we performed a feasibility experiment using a Robotic ENT (Ear, Nose, and Throat) Microsurgery System (REMS) which is a pre-clinical version developed by Galen Robotics (Baltimore, MD). The virtual models of REMS are loaded into AMBF and with their movements synchronized by referencing the actual state of the real REMS. The phantom was registered using optical tracking and the 3D Slicer registration toolkit.
Our system demonstrated efficient real-time tracking of the robotic system and anatomy, updating the virtual representations seamlessly in both AMBF and 3D Slicer.


\begin{figure}[t]
  \includegraphics[width=0.47\textwidth]{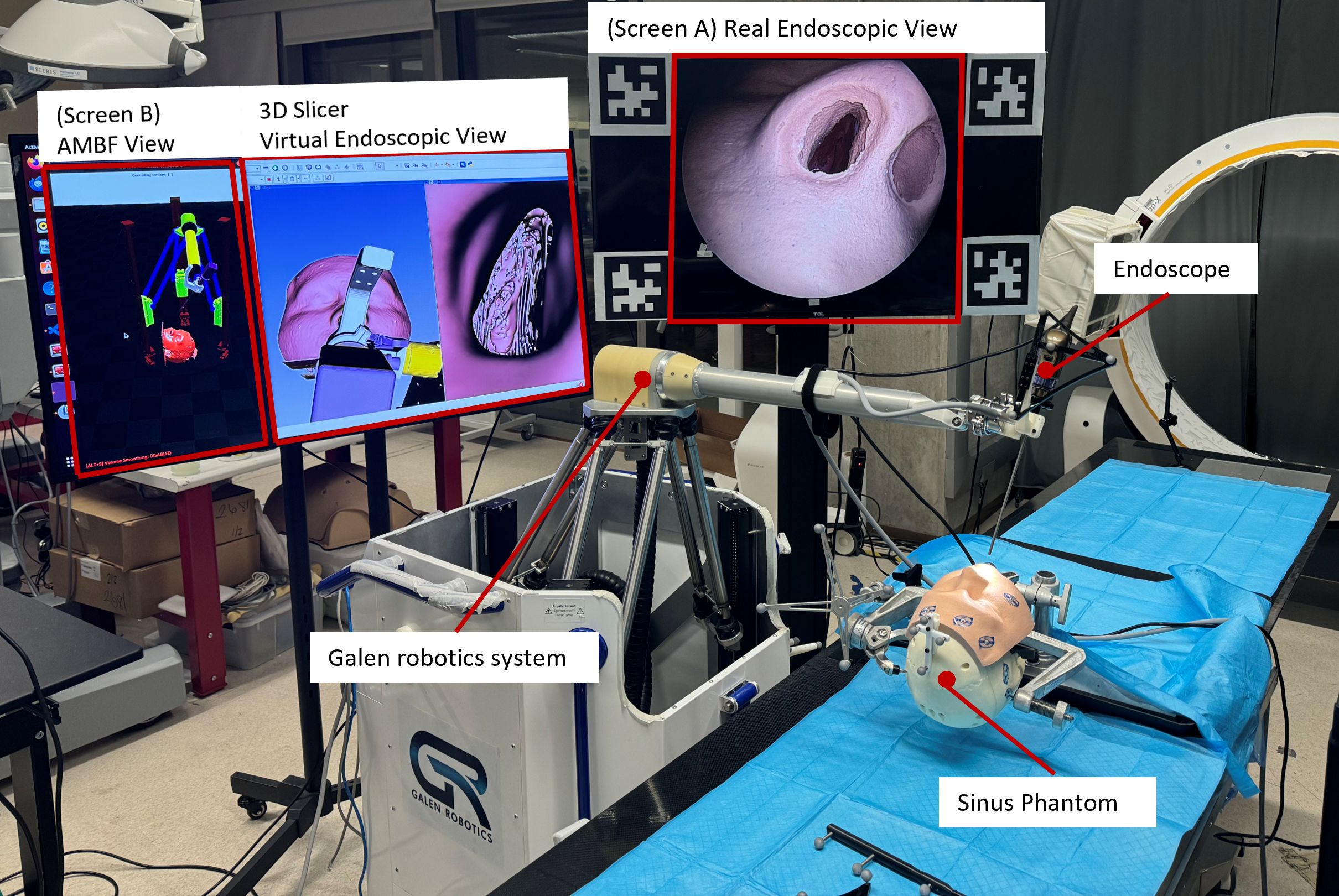}
  \centering
  \vspace{-0.3cm}
  \caption{Overview of the experimental setup for the simulated robotic sinus procedure. Galen robotics system holds an endoscope aimed at the sinus phantom (Screen A). AMBF simulator acquires motion data from the Galen robot, achieving real-time synchronization with the 3D Slicer application through the integration of a novel pipeline(Screen B).}
  \label{fig:model}
\end{figure}

\subsection{Performance Evaluation}
\begin{figure}[t]
  \includegraphics[width=0.47\textwidth]{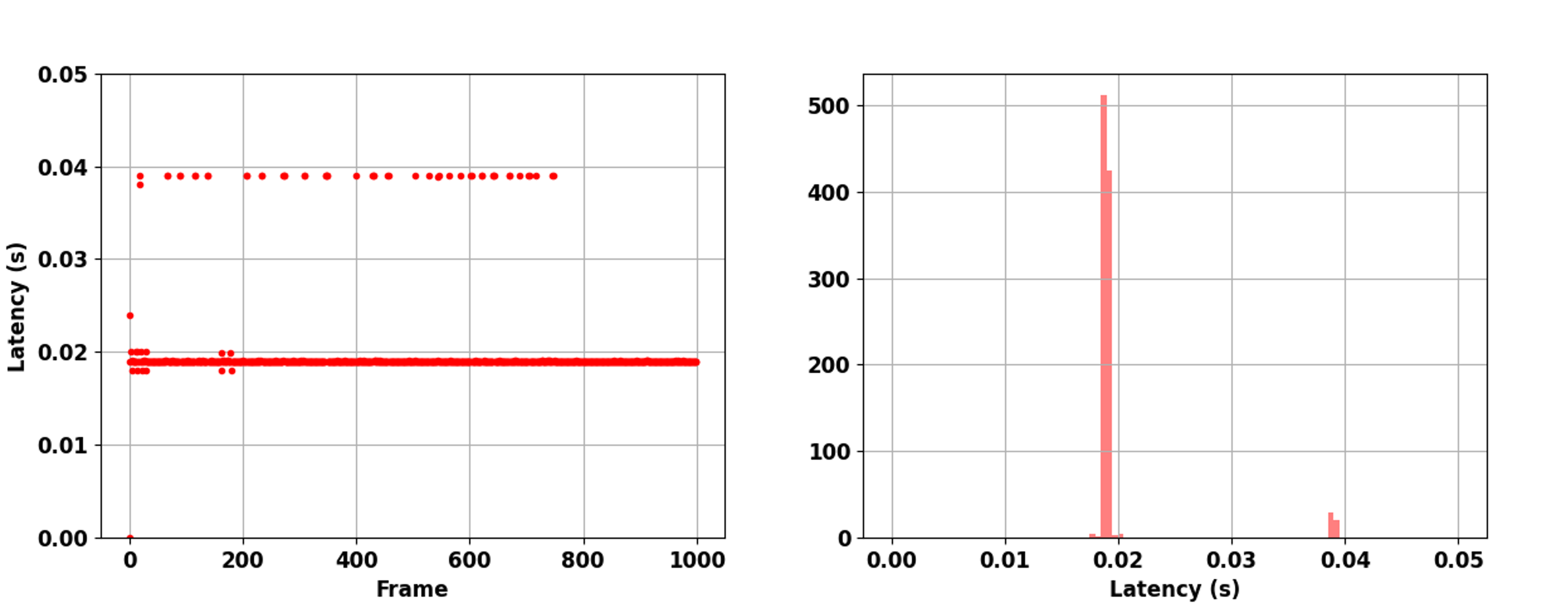}
  \centering
  \vspace{-0.3cm}
  \caption{Latency analysis: Round Time Delay(RTD) over 1000 frames}
  \label{fig:latency}
\end{figure}

For evaluating the performance of the system, the Galen Surgical Robot model comprising 25 rigid bodies and a skull model was loaded into AMBF. The 3D Slicer ROS module was designed to update at a maximum frequency of 200Hz. The evaluation consisted of actuating the robot within AMBF, and for each update, we measured the round-trip delay (RTD). The results are presented in Fig.\ref{fig:latency}, indicating \textbf{Mean:19.98\,ms, Median:18.99\,ms, std:4.40\,ms}. Given that our system focuses on the unilateral communication from AMBF to 3D Slicer, we can reasonably assume that there is approximately 10\,ms delay between the components. These statistics show that the RTD for the ROS module is stable at around 19\,ms (9.5\,ms for one-way delay).
Considering the lowest reported latency of 16\,ms that impacts human-computer interaction (HCI) tasks \cite{attig2017system}, we believe the latency of our system will not significantly affect the operator's tasks in real-world applications.

\section{Discussion \& Conclusion} 


In this work, we addressed an essential need for an open-source navigation and robotics platform integration by seamlessly integrating 3D Slicer, a widely adopted medical imaging platform, with the physics-based simulation environment AMBF, using ROS as the underlying communication middleware. The developed ROS modules, acting as nodes for both AMBF and 3D Slicer, serve as a bridge, facilitating real-time synchronization of scene states between the simulation and the 3D Slicer. Our implementation demonstrates the feasibility of achieving low-latency, real-time visualization of robotic interventions within the 3D Slicer environment.
This integration represents a substantial contribution to the field of robotic intervention research and development, harmonizing the strengths of 3D Slicer in visualization and registration with AMBF's dynamic and constraint-based simulation capabilities. The collaborative approach leverages ROS to enhance communication and holds the potential for future integration with additional open-source projects, fostering a more comprehensive toolset for researchers in medical robotics.


For future work, we aim to enhance the system by developing assistive shared control functions tailored for various robot-assisted ENT procedures. Recognizing surgeons' familiarity with the intuitive 3D Slicer interface, we plan to empower them with interactive planning within 3D Slicer, while incorporating assistive functions based on the physics interactions between the anatomy and surgical tools into AMBF. Computational feedback generated within AMBF will be dynamically reflected back to the 3D Slicer interface, offering surgeons a comprehensive and intuitive platform for planning intricate surgical procedures. This integration of computational feedback with the 3D Slicer interface aims to provide surgeons with a cohesive and user-friendly environment that aligns with their preferences and enhances their decision-making processes.

\section*{Acknowledgments and Disclosures}
This work was also supported in part by a research contract from Galen Robotics, by NIDCD K08 Grant DC019708, by a research agreement with the Hong Kong Multi-Scale Medical Robotics Centre,
and by Johns Hopkins University internal funds.
Russell Taylor and Johns Hopkins University (JHU) may be entitled to royalty payments related to the technology discussed in this paper, and Dr. Taylor has received or may receive some portion of these royalties. Also, Dr. Taylor is a paid consultant to and owns equity in Galen Robotics, Inc.  These arrangements have been reviewed and approved by JHU in accordance with its conflict of interest policy.



%
\bibliography{bibliography}

\begin{thebibliography}{10}
\providecommand{\url}[1]{#1}
\csname url@samestyle\endcsname
\providecommand{\newblock}{\relax}
\providecommand{\bibinfo}[2]{#2}
\providecommand{\BIBentrySTDinterwordspacing}{\spaceskip=0pt\relax}
\providecommand{\BIBentryALTinterwordstretchfactor}{4}
\providecommand{\BIBentryALTinterwordspacing}{\spaceskip=\fontdimen2\font plus
\BIBentryALTinterwordstretchfactor\fontdimen3\font minus \fontdimen4\font\relax}
\providecommand{\BIBforeignlanguage}[2]{{%
\expandafter\ifx\csname l@#1\endcsname\relax
\typeout{** WARNING: IEEEtran.bst: No hyphenation pattern has been}%
\typeout{** loaded for the language `#1'. Using the pattern for}%
\typeout{** the default language instead.}%
\else
\language=\csname l@#1\endcsname
\fi
#2}}
\providecommand{\BIBdecl}{\relax}
\BIBdecl

\bibitem{fichtinger2022image}
G.~Fichtinger, J.~Troccaz, and T.~Haidegger, ``Image-guided interventional robotics: Lost in translation?'' \emph{Proceedings of the IEEE}, vol. 110, no.~7, pp. 932--950, 2022.

\bibitem{cleary2010image}
K.~Cleary and T.~M. Peters, ``Image-guided interventions: technology review and clinical applications,'' \emph{Annual Review of Biomedical Engineering}, vol.~12, pp. 119--142, 2010.

\bibitem{taylor1999steady}
R.~Taylor, P.~Jensen \emph{et~al.}, ``A steady-hand robotic system for microsurgical augmentation,'' \emph{The International Journal of Robotics Research}, vol.~18, no.~12, pp. 1201--1210, 1999.

\bibitem{simaan2015intelligent}
N.~Simaan, R.~H. Taylor, and H.~Choset, ``Intelligent surgical robots with situational awareness,'' \emph{Mechanical Engineering}, vol. 137, no.~09, pp. S3--S6, 2015.

\bibitem{choi2021use}
H.~Choi, C.~Crump \emph{et~al.}, ``On the use of simulation in robotics: Opportunities, challenges, and suggestions for moving forward,'' \emph{PNAS}, vol. 118, no.~1, p. e1907856118, 2021.

\bibitem{fedorov20123dSlicer}
A.~Fedorov, R.~Beichel \emph{et~al.}, ``{3D Slicer} as an image computing platform for the quantitative imaging network,'' \emph{Magnetic Resonance Imaging}, vol.~30, no.~9, pp. 1323--1341, 2012.

\bibitem{ungi2016open}
T.~Ungi, A.~Lasso, and G.~Fichtinger, ``Open-source platforms for navigated image-guided interventions,'' \emph{Medical Image Analysis}, vol.~33, pp. 181--186, 2016.

\bibitem{quigley2009ros}
M.~Quigley, K.~Conley \emph{et~al.}, ``{ROS}: an open-source {Robot Operating System},'' in \emph{ICRA Workshop on Open Source Software}, vol.~3, no. 3.2.\hskip 1em plus 0.5em minus 0.4em\relax Kobe, Japan, 2009, p.~5.

\bibitem{connolly2021open}
L.~Connolly, A.~Deguet \emph{et~al.}, ``An open-source platform for cooperative, semi-autonomous robotic surgery,'' in \emph{IEEE International Conference on Autonomous Systems (ICAS)}.\hskip 1em plus 0.5em minus 0.4em\relax IEEE, 2021, pp. 1--5.

\bibitem{ambf}
A.~Munawar and G.~S. Fischer, ``An asynchronous multi-body simulation framework for real-time dynamics, haptics and learning with application to surgical robots.''\hskip 1em plus 0.5em minus 0.4em\relax IROS, Nov 2019.

\bibitem{slicerIGT}
T.~Ungi, A.~Lasso, and G.~Fichtinger, ``Open-source platforms for navigated image-guided interventions,'' \emph{Medical Image Analysis}, vol.~33, pp. 181--186, 2016.

\bibitem{tokuda2009openigtlink}
J.~Tokuda, G.~S. Fischer \emph{et~al.}, ``{OpenIGTLink}: an open network protocol for image-guided therapy environment,'' \emph{IJMRCAS}, vol.~5, no.~4, pp. 423--434, 2009.

\bibitem{lasso2014plus}
A.~Lasso, T.~Heffter \emph{et~al.}, ``{PLUS}: open-source toolkit for ultrasound-guided intervention systems,'' \emph{TBME}, vol.~61, no.~10, pp. 2527--2537, 2014.

\bibitem{tauscher2015openigtlink}
S.~Tauscher, J.~Tokuda \emph{et~al.}, ``{OpenIGTLink} interface for state control and visualisation of a robot for image-guided therapy systems,'' \emph{IJCARS}, vol.~10, pp. 285--292, 2015.

\bibitem{frank2017ros}
T.~Frank, A.~Krieger \emph{et~al.}, ``{ROS-IGTL-Bridge}: an open network interface for image-guided therapy using the {ROS} environment,'' \emph{IJCARS}, vol.~12, pp. 1451--1460, 2017.

\bibitem{ros2_module_connolly}
L.~Connolly \emph{et~al.}, ``Bridging {3D Slicer} and {ROS2} for image-guided robotic interventions.'' \emph{Sensors}, vol. 22,14 5336, Jul 2022.

\bibitem{foote2013tf}
T.~Foote, ``tf: The transform library,'' in \emph{TePRA}.\hskip 1em plus 0.5em minus 0.4em\relax IEEE, 2013, pp. 1--6.

\bibitem{koenig2004gazebo}
N.~Koenig and A.~Howard, ``Design and use paradigms for gazebo, an open-source multi-robot simulator,'' in \emph{IROS}, vol.~3.\hskip 1em plus 0.5em minus 0.4em\relax IEEE, 2004, pp. 2149--2154.

\bibitem{rohmer2013vrep}
E.~Rohmer, S.~P. Singh, and M.~Freese, ``{V-REP}: A versatile and scalable robot simulation framework,'' in \emph{IEEE/RSJ International Conference on Intelligent Robots and Systems (IROS)}.\hskip 1em plus 0.5em minus 0.4em\relax IEEE, 2013, pp. 1321--1326.

\bibitem{munawar2023fully}
A.~Munawar, Z.~Li \emph{et~al.}, ``Fully immersive virtual reality for skull-base surgery: surgical training and beyond,'' \emph{IJCARS}, pp. 1--9, 2023.

\bibitem{phalen2023platform}
H.~Phalen, A.~Munawar \emph{et~al.}, ``Platform for investigating continuum manipulator behavior in orthopedics,'' \emph{IJCARS}, pp. 1--6, 2023.

\bibitem{shu2023twin}
H.~Shu, R.~Liang \emph{et~al.}, ``Twin-s: a digital twin for skull base surgery,'' \emph{IJCARS}, pp. 1--8, 2023.

\bibitem{munawar2022open}
A.~Munawar, J.~Y. Wu \emph{et~al.}, ``Open simulation environment for learning and practice of robot-assisted surgical suturing,'' \emph{RAL}, vol.~7, no.~2, pp. 3843--3850, 2022.

\bibitem{attig2017system}
C.~Attig, N.~Rauh \emph{et~al.}, ``System latency guidelines then and now--is zero latency really considered necessary?'' in \emph{Engineering Psychology and Cognitive Ergonomics}.\hskip 1em plus 0.5em minus 0.4em\relax Springer, 2017, pp. 3--14.

\end{thebibliography}
\bibliographystyle{IEEEtran}

\end{document}